\documentclass[runningheads]{llncs}
\usepackage{graphicx}

\usepackage{tikz}
\usepackage{comment}
\usepackage{amsmath,amssymb} 
\usepackage{color}
\usepackage{multirow}

\usepackage[accsupp]{axessibility}  

\begin{document}

\pagestyle{headings}
\mainmatter
\def\ECCVSubNumber{15}  

\title{Multi-Task Learning Framework for \\Emotion Recognition in-the-wild} 

\titlerunning{Multi-Task Learning Framework for Emotion Recognition in-the-wild}
\authorrunning{Tenggan Zhang, Chuanhe Liu, Xiaolong Liu, et al.} 
\author{Anonymous ECCV submission}
\institute{Paper ID \ECCVSubNumber}

\author{Tenggan Zhang\inst{1,\ast} \and Chuanhe Liu\inst{2,\ast} \and Xiaolong Liu\inst{2,\ast} \and Yuchen Liu\inst{1} \and
Liyu Meng\inst{2} \and Lei Sun\inst{1} \and 
Wenqiang Jiang\inst{2} \and Fengyuan Zhang\inst{1} \and 
Jinming Zhao\inst{3} \and Qin Jin\inst{1,\dagger}}
\institute{School of Information, Renmin University of China\\ 
\email{\{zhangtenggan,zbxytx,qjin\}@ruc.edu.cn} \\
\email{\{sunlei.ruc,zhangfy.ruc\}@gmail.com} \and
Beijing Seek Truth Data Technology Co.,Ltd.\\
\email{\{liuchuanhe,liuxiaolong,mengliyu,jiangwenqiang\}@situdata.com} \and
Qiyuan Lab, Beijing, China\\
\email{zhaojinming@qiyuanlab.com}}

\def\thefootnote{$\ast$}\footnotetext{Equal Contribution.}
\def\thefootnote{$\dagger$}\footnotetext{Corresponding Author.}

\maketitle

\begin{abstract}

This paper presents our system for the Multi-Task Learning (MTL) Challenge in the 4th Affective Behavior Analysis in-the-wild (ABAW) competition. 
We explore the research problems of this challenge from three aspects: 1) For obtaining efficient and robust visual feature representations, we propose MAE-based unsupervised representation learning and IResNet/DenseNet-based supervised representation learning methods;
2) Considering the importance of temporal information in videos, we explore three types of sequential encoders to capture the temporal information, including the encoder based on transformer, the encoder based on LSTM, and the encoder based on GRU; 
3) For modeling the correlation between these different tasks (i.e., valence, arousal, expression, and AU) for multi-task affective analysis, we first explore the dependency between these different tasks and propose three multi-task learning frameworks to model the correlations effectively.
Our system achieves the performance of $1.7607$ on the validation dataset and $1.4361$ on the test dataset, ranking first in the MTL Challenge.
The code is available at https://github.com/AIM3-RUC/ABAW4.

\end{abstract}

\section{Introduction}

Affective computing aims to develop technologies to empower machines with the capability of observing, interpreting, and generating emotions just like humans do~\cite{DBLP:conf/acii/TaoT05}. There has emerged a wide range of application scenarios of affective computing, including health research, society analysis, and other interaction scenarios. 
More and more people are interested in affective computing due to the significant improvement of machine learning technology performance and the growing attention to the mental health field. There are lots of datasets to support the research of affective computing, including Aff-wild~\cite{kollias2017recognition}, Aff-wild2~\cite{AffWild2}, and s-Aff-Wild2\cite{kollias2022abaw4,kollias2022abaw,kollias2021distribution,kollias2021affect,kollias2020deep,kollias2020va,kollias2019expression,kollias2019deep,kollias2018photorealistic,zafeiriou2017aff,kollias2017recognition}. The advancement of multi-task learning algorithms~\cite{DBLP:journals/corr/Ruder17a} 
has also boosted performance via exploring supervision from different tasks. 

Our system for the Multi-Task Learning (MTL) Challenge contains four key components. 
1) We explore several unsupervised (MAE-based) and supervised (IResNet/DenseNet-based) visual feature representation learning methods for learning effective and robust visual representations; 
2) We utilize three types of temporal encoders, including GRU~\cite{chung2014empirical}, LSTM~\cite{sak2014long} and Transformer~\cite{vaswani2017attention}, to capture the sequential information in videos; 
3) We employ multi-task frameworks to predict the valence, arousal, expression and AU values. Specifically, we investigate three different strategies for multi-task learning, namely \underline{S}hare \underline{E}ncoder (SE), \underline{S}hare \underline{B}ottom of \underline{E}ncoder (SBE) and \underline{S}hare \underline{B}ottom of \underline{E}ncoder with \underline{H}idden \underline{S}tates \underline{F}eedback (SBE-HSF); 
4) Finally, we adopt ensemble strategies and cross-validation to further enhance the predictions, and we get the performance of $1.7607$ on the validation dataset and $1.4361$ on the test dataset, ranking first in the MTL Challenge. 
\section{Related Works}
There are lots of solutions proposed for former ABAW competitions. We investigate some studies for valence and arousal prediction, facial expression classification and facial action unit detection, which are based on deep learning methods. 

For valence and arousal prediction, \cite{meng2022valence} proposes a novel architecture to fuse temporal-aware multimodal features and an ensemble method to further enhance performance of regression models. \cite{zhang2022continuous} proposes a model for continuous emotion prediction using a cross-modal co-attention mechanism with three modalities (i.e., visual, audio and linguistic information). \cite{nguyen2022ensemble} combines local attention with GRU and uses multimodal features to enhance the performance.
For expression classification, facing the problem that the changes of features for expression are difficult to be processed by one attention module, \cite{wen2021distract} proposes a novel attention mechanism to capture local and semantic features. \cite{zhang2022transformer} utilizes multimodal features, including visual, audio and text to build a transformer-based framework for expression classification and AU detection. 
For facial action unit detection, \cite{jacob2021facial} utilizes a multi-task approach with a center contrastive loss and ROI attention module to learn the correlations of facial action units. \cite{jiang2022model} proposes a model-level ensemble method to achieve comparable results. \cite{fan2020facial} introduces a semantic correspondence convolution module to capture the relations of AU in a heat map regression framework dynamically. 

\section{Method}

\begin{figure*}[t]
\centering
\includegraphics[width=1\textwidth]{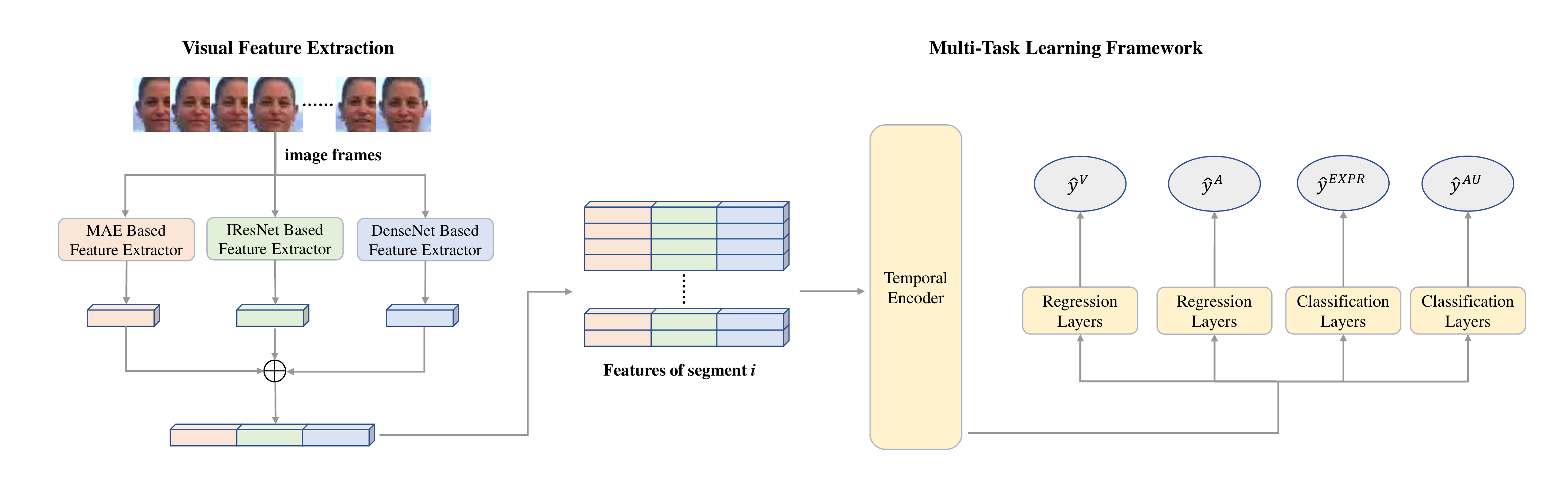}
\caption{The pipeline of our method for the challenge. }
\label{fig:overall}
\end{figure*}

Given an image sequence consisting of $\{F_{1}, F_{2}, ..., F_{n}\}$ from video $X$, the goal of the MTL challenge is to produce four types of emotion predictions for each frame, including the label $y^{v}$ for valence, the label $y^{a}$ for arousal, the label $y^{e}$ for expression, and the labels $\{y^{AU1}, y^{AU2}, ..., y^{AU26}\}$ for 12 AUs. Please note that only some sampled frames in a video are annotated in the training data, and the four types of annotations may be partially missing for an image frame. 
Our pipeline for the challenge is shown in figure \ref{fig:overall}. 

\subsection{Features}

\subsubsection{MAE-based Features}
The features of the first type are extracted by MAE~\cite{MAE} models\footnote{https://github.com/pengzhiliang/MAE-pytorch} which use C-MS-Celeb~\cite{jin2018community} and EmotionNet~\cite{emotionnet} datasets at the pre-training stage. 
The first model is pre-trained on the C-MS-Celeb dataset and fine-tuned on different downstream tasks, including expression classification on the s-Aff-Wild2 dataset, AU classification task on the s-Aff-Wild2 dataset, expression classification on the AffectNet~\cite{mollahosseini2017affectnet} dataset and expression classification on the dataset combining FER+~\cite{BarsoumICMI2016} and AffectNet~\cite{mollahosseini2017affectnet} datasets. 
As for the second model, we first use the EmotionNet dataset to pre-train the MAE model with the reconstruction task, and then use the AffectNet~\cite{mollahosseini2017affectnet} dataset to fine-tune the model further. 

\subsubsection{IResNet-based Features}
The features of the second type are extracted by IResNet100 models. The models are pre-trained in two different settings. 
As for the first setting, we use FER+~\cite{BarsoumICMI2016}, RAF-DB~\cite{li2017reliable,li2019reliable}, and AffectNet~\cite{mollahosseini2017affectnet} datasets to pre-train the model. Specifically, the faces are aligned by keypoints and the input size is resized into 112x112 before pre-training. As for the second setting, we use the Glint360K~\cite{an2021partial} dataset to pre-train the model, and then use an FAU dataset with commercial authorization to train this model further. 

\subsubsection{DenseNet-based Features}
The features of the third type are extracted by a DenseNet~\cite{iandola2014densenet} model. The pre-training stage uses FER+ and AffectNet datasets, and we also try to fine-tune the pre-trained model on the s-Aff-Wild2 dataset, including the expression classification task and AU classification task.

\subsection{Temporal Encoder}
Because the GPU memory is limited, the annotated frames are firstly split into segments. If the length of the split segment is $l$ and $n$ available annotated frames are contained in the video, we can split the frames into $[n/l]+1$ segments, which means annotated frames $\{F_{(i-1)*l+1}, ..., F_{(i-1)*l+l}\}$ are contained in the $i$-th segment. 
After getting the visual features from the $i$-th segment $f^{m}_{i}$, three different temporal encoders including GRU, LSTM and transformer encoder are used to capture the temporal information in the video.

\subsubsection{GRU-based Temporal Encoder}
We use a Gate Recurrent Unit Network (GRU) to encode the temporal information of the image sequence. Segment $s_{i}$ means the $i$-th segment, and $f^{m}_{i}$ means the input of GRU is the visual features for $s_{i}$. Furthermore, the hidden states of the last layer are fed from the previous segment $s_{i-1}$ into the GRU to utilize the information from the last segment.
\begin{small}
\begin{equation}
    \centering
    g_{i}, h_{i} = \text{GRU}(f^{m}_{i}, h_{i-1})
\end{equation}
\end{small}

where $h_{i}$ denotes the hidden states at the end of $s_{i}$. $h_{0}$ is initialized to be zeros. To ensure that the last frame of $s_{i-1}$ and the first frame of segment $s_{i}$ are consecutive frames, there is no overlap between the two adjacent segments. 

\subsubsection{LSTM-based Temporal Encoder}
We employ a Long Short-Term Memory Network (LSTM) to model the sequential dependencies in the video. It can be formulated as follows:
\begin{small}
\begin{equation}
    \centering
    g_{i}, h_{i} = \text{LSTM}(f^{m}_{i}, h_{i-1})
\end{equation}
\end{small}

The symbols have the same meaning as in the GRU part. 

\subsubsection{Transformer-based Temporal Encoder}
We utilize a transformer encoder to model the temporal information in the video segment as well, which can be formulated as follows:

\begin{small}
\begin{equation}
    \centering
    g_{i} = \text{TRMEncoder}(f^{m}_{i})
\end{equation}
\end{small}

Unlike GRU and LSTM, the transformer encoder just models the context in a single segment and ignores the dependencies of frames between segments.

\subsection{Single Task Loss Function}

We first introduce the loss function for each task in this subsection. 

\vspace{10pt}
\noindent\textbf{\textit{{Valence and Arousal estimation task}}}: 

We utilize the Mean Squared Error (MSE) loss which can be formulated as 

\begin{small}
    \begin{equation}
        \centering
        L^{V} = L^{A} = \frac{1}{N}\sum_{i=1}^{N}(y_{i}-\hat{y}_{i})
        \end{equation}
\end{small}
where $N$ denotes the number of frames in each batch, $\hat{y}_{i}$ and $y_{i}$ denote the prediction and label of valence or arousal in each batch respectively. 

\vspace{10pt}
\noindent\textbf{\textit{{Expression Classification task}}}:

We utilize the Cross Entropy (CE) loss which can be formulated as

\begin{small}
    \begin{equation}
        \centering
        L^{EXPR} = -\sum_{i=1}^{N}\sum_{j=1}^{C}y_{ij}log(\hat{y}_{ij})
        \end{equation}
\end{small}
where $C$ is equal to $8$ which denotes the total classification number of all expression, $\hat{y}_{ij}$ and $y_{ij}$ denote the prediction and label of expression in each batch.

\vspace{10pt}
\noindent\textbf{\textit{{AU Classification task}}}:

We utilize Binary Cross Entropy (BCE) loss which can be formulated as

\begin{small}
    \begin{equation}
        \centering
        L^{AU} = \sum_{i=1}^{N}\sum_{j=1}^{M}(-(y_{ij}log(\hat{y}_{ij} + (1 - y_{ij}) log(1 - \hat{y}_{ij}))))
        \end{equation}
\end{small}
where $M$ is equal to $12$ which denotes the total number of facial action units, $\hat{y}_{ij}$ and $y_{ij}$ denote the logits and label of facial action units in each batch.

\subsection{Multi-Task Learning Framework}
As we mentioned above, the overall estimation objectives can be divided into four tasks, including the estimation of valence, arousal, expression and action units on expressive facial images. 
These four objectives focus on different information on the facial images, where the essential information about one task may be helpful to the modeling of some other tasks. 

The dependencies between tasks are manifested mainly in two aspects: 
First, the low-level representations are common for some tasks and they can be shared to benefit each task. 
Second, some high-level task-specific information of one task could be important features for other tasks. For example, since the definition of expressions depends on facial action units to some extent, the high-level features in the AU detection task can help the estimation of expression.

\begin{figure*}[t]
\centering
\includegraphics[width=1\textwidth]{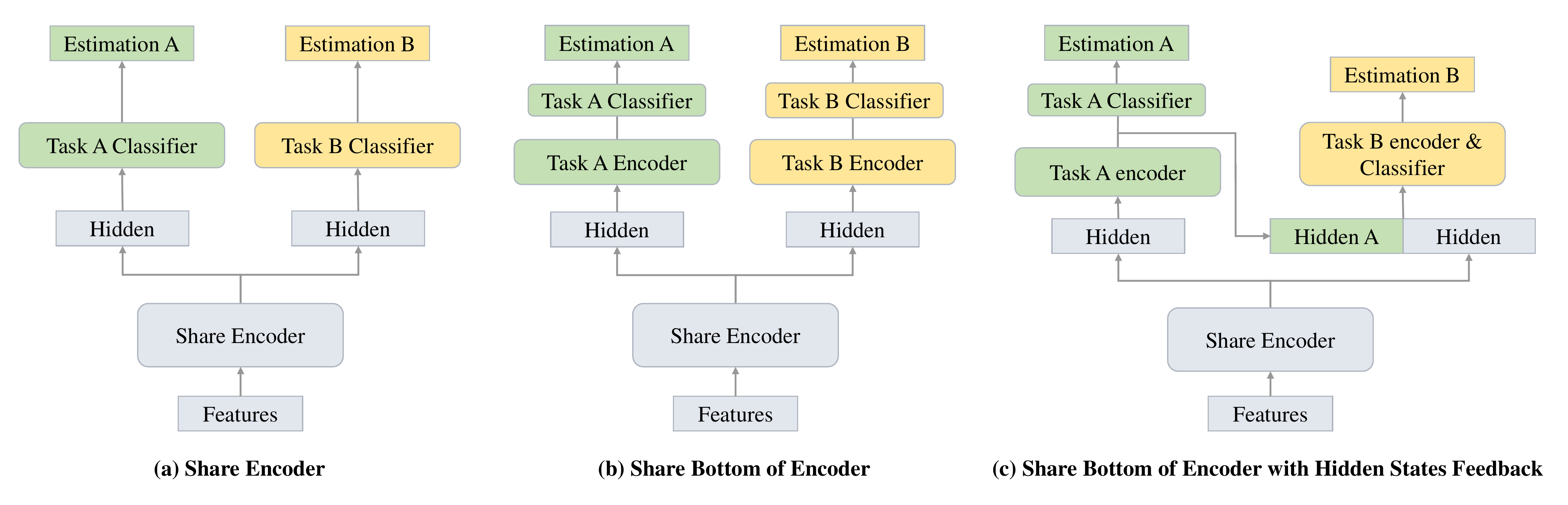}
\caption{Our proposed multi-task learning frameworks.}
\label{fig:mtl}
\end{figure*}

In order to make use of such dependencies between different tasks, we make some efforts on the multi-task learning frameworks instead of the single-task models. 
Specifically, we propose three multi-task learning frameworks, as illustrated in Figure \ref{fig:mtl}.

\subsubsection{Share Encoder}
We propose the \underline{S}hare \underline{E}ncoder (SE) framework as the baseline, which is commonly used in the field of multi-task learning. In the SE framework, the temporal encoder is directly shared between different tasks, while each task retains task-specific regression or classification layers. The structure of the SE framework is shown in Figure \ref{fig:mtl}(a), which can be formulated as follows:
\begin{small}
\begin{align}
    \centering
    g_{i} &= \text{TE}(f^{m}_{i}) \\
    \hat{y}^{t}_{i} &= W^{t}_{p}g_{i} + b^{t}_{p},\ t \in T
\end{align}
\end{small}
where TE denotes the temporal encoder, $T$ denotes the collection of chosen tasks in the multi-task learning framework, $t$ denotes a specific task in \{v, a, e, au\}, ${y}^{t}_{i}$ denotes the predictions of task $t$ of segment $s_{i}$, $W^{t}_{p}$ and $b^{t}_{p}$ denote the parameters to be optimized.

\subsubsection{Share Bottom of Encoder}
Under the assumption that the bottom layers of the encoder capture more basic information in facial images while the top layers encode more task-specific features, we propose to only share the bottom layers of the temporal encoder between different tasks. The structure of the \underline{S}hare \underline{B}ottom of \underline{E}ncoder (SBE) framework is shown in Figure \ref{fig:mtl}(b), which can be formulated as follows:
\begin{small}
\begin{align}
    \centering
    g_{i} &= \text{TE}(f^{m}_{i}) \\
    g^{t}_{i} &= \text{TE}^{t}(g_{i}),\ t \in T \\
    \hat{y}^{t}_{i} &= W^{t}_{p}g^{t}_{i} + b^{t}_{p},\ t \in T
\end{align}
\end{small}
where TE denotes the temporal encoder, $t$ denotes a specific task and $T$ denotes the collection of chosen tasks, $TE^{t}$ denotes the task-specif temporal encoder of task $t$, ${y}^{t}_{i}$ denotes the predictions of task $t$ of segment $s_{i}$, $W^{t}_{p}$ and $b^{t}_{p}$ denote the parameters to be optimized.

\subsubsection{Share Bottom of Encoder with Hidden States Feedback}
Although the proposed SBE framework has captured the low-level shared information between different tasks, it might ignore the high-level task-specific dependencies of tasks. In order to model such high-level dependencies, we propose the \underline{S}hare \underline{B}ottom of \underline{E}ncoder with \underline{H}idden \underline{S}tates \underline{F}eedback (SBE-HSF) framework, as illustrated in Figure \ref{fig:mtl}(c). In the SBE-HSF framework, all the tasks share the bottom layers of the temporal encoder and retain task-specific top layers, as in the SBE framework. 

Afterward, considering that the information of one task could benefit the estimation of another task, we feed the last hidden states of the temporal encoder of the source task into the temporal encoder of the target task as features. It can be formulated as follows:
\begin{small}
\begin{align}
    \centering
    g_{i} &= \text{TE}(f^{m}_{i}) \\
    g^{t}_{i} &= \text{TE}^{t}(g_{i}),\ t \in T\setminus\{t^{tgt}\} \\
    g^{tgt}_{i} &= \text{TE}^{tgt}(\text{Concat}(g_{i}, g^{src}_{i})) \\
    \hat{y}^{t}_{i} &= W^{t}_{p}g^{t}_{i} + b^{t}_{p},\ t \in T
\end{align}
\end{small}
where TE denotes the temporal encoder, $t$ denotes a specific task and $T$ denotes the collection of chosen tasks, $src$ and $tgt$ denote the source and target task of the feedback structure, respectively, $TE^{t}$ denotes the task-specif temporal encoder of task $t$, ${y}^{t}_{i}$ denotes the predictions of task $t$ of segment $s_{i}$, $W^{t}_{p}$ and $b^{t}_{p}$ denote the parameters to be optimized. In addition, in the backward propagation stage, the gradient of $g^{src}_{i}$ is detached.

\subsubsection{Multi-Task Loss Function}
In the multi-task learning framework, we utilize the multi-task loss function to optimize the model, which combines the loss functions of all tasks chosen for multi-task learning:
\begin{small}
    \begin{equation}
        \centering
        L = \sum_{t \in T} \alpha^{t} L^{t}
        \label{equal:13}
        \end{equation}
\end{small}
where $t$ denotes a specific task and $T$ denotes the collection of chosen tasks,  $L^{t}$ denotes the loss function of task $t$, which is mentioned above, $\alpha^{t}$ denotes the weight of $L^{t}$ which is a hyper-parameter.
\section{Experiments}\label{expriemnts}
\subsection{Dataset}
The Multi-Task Learning (MTL) Challenge in the fourth ABAW competition\cite{kollias2022abaw4} uses the s-Aff-Wild2 dataset as the competition corpora, which is the static version of the Aff-Wild2\cite{AffWild2} database and contains some specific frames of the Aff-Wild2 database. 

As for feature extractors, the FER+\cite{BarsoumICMI2016}, RAF-DB\cite{li2017reliable,li2019reliable}, AffectNet\cite{mollahosseini2017affectnet}, C-MS-Celeb\cite{jin2018community} and EmotionNet\cite{emotionnet} datasets are used for pre-training. 
In addition, an authorized commercial FAU dataset is also used to pre-train the visual feature extractor. It contains 7K images in 15 face action unit categories(AU1, AU2, AU4, AU5, AU6, AU7, AU9, AU10, AU11, AU12, AU15, AU17, AU20, AU24, and AU26).

\subsection{Experiment Setup}
As for the training setting, we use Nvidia GeForce GTX 1080 Ti GPUs to train the models, and the optimizer is the Adam\cite{kingma2014adam}.
The number of epochs is 50, the dropout rate of the temporal encoder and the FC layers is 0.3, the learning rate is 0.00005, the length of video segments is 250 for arousal and 64 for valence, expression and AU, and the batch size is 8. 

As for the model architecture, the dimension of the feed-forward layers or the size of hidden states is 1024, the number of FC layers is 3 and the sizes of hidden states are \{512, 256\}. Specially, the encoder of transformer has 4 layers and 4 attention heads. 

As for the smooth strategy, we search for the best window of valence and arousal for each result based on the performance on the validation set. Most window lengths are 5 and 10.

\subsection{Overall Results on the validation set}
In this section, we will demonstrate the overall experimental results of our proposed method for the valence, arousal, expression and action unit estimation tasks. Specifically, the experimental results are divided into three parts, including the single-task results, the exploration of multi-task dependencies and the results of multi-task learning frameworks. We report the average performance of 3 runs with different random seeds. 

\subsubsection{Single-Task Results}
In order to verify the performance of our proposed model without utilizing the multi-task dependencies, we conduct several single-task experiments. The results are demonstrated in Table \ref{tab:single}. 
\begin{table}[]
\begin{center}
  \caption{The performance of our proposed method on the validation set for each single task.}
  \label{tab:single}
\begin{tabular}{c|c|c|c}
\hline
Model       & Task       & Features                 & Performance \\ \hline
Transformer & Valence    & MAE,ires100,fau,DenseNet & 0.6414      \\
Transformer & Arousal    & MAE,ires100,fau,DenseNet & 0.6053      \\
Transformer & EXPR       & MAE,ires100,fau,DenseNet & 0.4310      \\
Transformer & AU         & MAE,ires100,fau,DenseNet & 0.4994      \\ \hline
\end{tabular}
\end{center}
\end{table}

\subsubsection{Results of Multi-Task Learning Frameworks}
We try different task combinations and apply the best task combination to the multi-task learning frameworks for each task. As a result, we find the best task combinations as follows: \{V, EXPR\} for valence, \{V, A, AU\} for arousal, \{V, EXPR\} for expression and \{V, AU\} for action unit.
The experimental results of our proposed multi-task learning frameworks and the comparison with single-task models are shown in Table \ref{tab:mtl}. Specifically, the combination of features is the same as that in single-task settings, and the set of tasks chosen for the multi-task learning frameworks is based on the multi-task dependencies, which have been explored above.

As is shown in the table, first, all of our proposed multi-task frameworks outperform the single-task models on valence, expression and action unit estimation tasks. On the arousal estimation task, only the SE framework performs inferior to the single-task model and the other two frameworks outperform it. These results show that our proposed multi-task learning frameworks can improve performance and surpass the single-task models.

Moreover, the two proposed frameworks, SBE and SBE-HSF, show the advanced performance, where the former is an improvement on the SE framework and the latter is an improvement on the former. The SBE framework outperforms the SE frameworks, and the SBE-HSF framework outperforms the SBE framework on arousal, expression and action unit estimation tasks. It indicates our proposed multi-task learning framework can effectively improve performance.

\begin{table}[]
\caption{The performance of our proposed multi-task learning frameworks on the validation set.}
\resizebox{\columnwidth}{!}{
\label{tab:mtl}
\begin{tabular}{c|clc|clc|clc|clc}
\hline
\multirow{2}{*}{}         & \multicolumn{3}{c|}{Valence}                             & \multicolumn{3}{c|}{Arousal}                                      & \multicolumn{3}{c|}{EXPR}                                   & \multicolumn{3}{c}{AU}                                          \\ \cline{2-13} 
                          & \multicolumn{2}{c|}{Tasks}     & CCC                     & \multicolumn{2}{c|}{Tasks}     & CCC                              & \multicolumn{2}{c|}{Tasks}     & F1                               & \multicolumn{2}{c|}{Tasks}   & F1                               \\ \hline
Single Task               & \multicolumn{2}{c|}{V}         & 0.6414                  & \multicolumn{2}{c|}{A}         & 0.6053                           & \multicolumn{2}{c|}{EXPR}      & 0.4310                           & \multicolumn{2}{c|}{AU}      & 0.4994                           \\ \hline
SE                     & \multicolumn{2}{c|}{V, EXPR}   & 0.6529                  & \multicolumn{2}{c|}{V, A, AU}  & 0.5989                           & \multicolumn{2}{c|}{V, EXPR}   & 0.4406                           & \multicolumn{2}{c|}{V, AU}   & 0.5084                           \\ \hline
SBE              & \multicolumn{2}{c|}{V, EXPR}   & \textbf{0.6558}         & \multicolumn{2}{c|}{V, A, AU}  & 0.6091                           & \multicolumn{2}{c|}{V, EXPR}   & 0.4460                           & \multicolumn{2}{c|}{V, AU}   & 0.5107                           \\ \hline
\multirow{2}{*}{SBE-HSF} & \multicolumn{2}{c|}{Src: V}    & \multirow{2}{*}{0.6535} & \multicolumn{2}{c|}{Src: V,AU} & \multirow{2}{*}{\textbf{0.6138}} & \multicolumn{2}{c|}{Src: EXPR} & \multirow{2}{*}{\textbf{0.4543}} & \multicolumn{2}{c|}{Src: V}  & \multirow{2}{*}{\textbf{0.5138}} \\ \cline{2-3} \cline{5-6} \cline{8-9} \cline{11-12}
                          & \multicolumn{2}{c|}{Tgt: EXPR} &                         & \multicolumn{2}{c|}{Tgt: A}    &                                  & \multicolumn{2}{c|}{Tgt: V}    &                                  & \multicolumn{2}{c|}{Tgt: AU} &                                  \\ \hline
\end{tabular}}
\end{table}

\subsection{Model Ensemble}

\begin{table}[]
\centering
  \caption{The single model results and ensemble result on the validation set for the valence prediction task.}
  \label{tab:v}
\begin{tabular}{cccc}
\hline
\multicolumn{1}{c|}{Model}       & \multicolumn{1}{c|}{Features}                 & \multicolumn{1}{c|}{Loss}   & Valence-CCC \\ \hline
\multicolumn{1}{c|}{Transformer} & \multicolumn{1}{c|}{MAE,ires100,fau,DenseNet} & \multicolumn{1}{c|}{V,EXPR} & 0.6778      \\
\multicolumn{1}{c|}{LSTM}        & \multicolumn{1}{c|}{MAE,ires100,fau,DenseNet} & \multicolumn{1}{c|}{V}      & 0.6734      \\ \hline
\textbf{Ensemble} &  &  & \textbf{0.7101} \\ \hline
\end{tabular}
\end{table}

\begin{table}[]
\centering
  \caption{The single model results and ensemble result on the validation set for the arousal prediction task.}
  \label{tab:a}
\begin{tabular}{cccc}
\hline
\multicolumn{1}{c|}{Model} & \multicolumn{1}{c|}{Features}                 & \multicolumn{1}{c|}{Loss}   & Arousal-CCC     \\ \hline
\multicolumn{1}{c|}{LSTM}  & \multicolumn{1}{c|}{MAE,ires100,fau,DenseNet} & \multicolumn{1}{c|}{V,A,AU} & 0.6384          \\
\multicolumn{1}{c|}{LSTM}  & \multicolumn{1}{c|}{MAE,ires100,fau,DenseNet} & \multicolumn{1}{c|}{V,A,AU} & 0.6354          \\
\multicolumn{1}{c|}{GRU}   & \multicolumn{1}{c|}{MAE,ires100,fau,DenseNet} & \multicolumn{1}{c|}{V,A,AU} & 0.6292          \\
\multicolumn{1}{c|}{GRU}   & \multicolumn{1}{c|}{MAE,ires100,DenseNet}     & \multicolumn{1}{c|}{V,A,AU} & 0.6244          \\ \hline
\textbf{Ensemble}          &                                               &                             & \textbf{0.6604} \\ \hline
\end{tabular}
\end{table}

 We evaluate the proposed methods for the valence and arousal prediction task on the validation set. As is shown in the Table \ref{tab:v} and Table \ref{tab:a}, the best performance for valence is achieved by transformer-based model, and the best performance for arousal is achieved by LSTM-based model and the GRU-based model also achieves competitive performance for arousal. Furthermore, the ensemble result can achieve 0.7101 on valence and 0.6604 on arousal, which shows that the results of different models benefit each other.

\begin{table}[]
\centering
  \caption{The single model results and ensemble result on the validation set for the EXPR prediction task.}
  \label{tab:expr}
\begin{tabular}{cccc}
\hline
\multicolumn{1}{c|}{Model}       & \multicolumn{1}{c|}{Features}                 & \multicolumn{1}{c|}{Loss}   & EXPR-F1         \\ \hline
\multicolumn{1}{c|}{Transformer} & \multicolumn{1}{c|}{MAE,ires100,fau,DenseNet} & \multicolumn{1}{c|}{V,EXPR} & 0.4739          \\
\multicolumn{1}{c|}{Transformer} & \multicolumn{1}{c|}{MAE,ires100,fau,DenseNet} & \multicolumn{1}{c|}{V,EXPR} & 0.4796          \\ \hline
\textbf{Ensemble}                &                                               &                             & \textbf{0.5090} \\ \hline
\end{tabular}
\end{table}

Table \ref{tab:expr} shows the results on the validation set for expression prediction. As is shown in the table, the transformer-based model can achieve the best performance for expression and the ensemble result can achieve 0.5090 on the validation set. We use the vote strategy for expression ensemble, and we choose the class with the least number in the training set when the number of classes with the most votes is more than one. 

\begin{table}[]
\centering
  \caption{The single model results and ensemble result on the validation set for the AU prediction task.}
  \label{tab:AU}
\begin{tabular}{ccccc}
\hline
\multicolumn{1}{c|}{Model}         & \multicolumn{1}{c|}{Features}                 & \multicolumn{1}{c|}{Loss} & \multicolumn{1}{c|}{Threshold} & AU-F1           \\ \hline
\multicolumn{1}{c|}{Transformer}   & \multicolumn{1}{c|}{MAE,ires100,fau,DenseNet} & \multicolumn{1}{c|}{V,AU} & \multicolumn{1}{c|}{0.5}       & 0.5217          \\
\multicolumn{1}{c|}{Transformer}   & \multicolumn{1}{c|}{MAE,ires100,fau,DenseNet} & \multicolumn{1}{c|}{V,AU} & \multicolumn{1}{c|}{0.5}       & 0.5213          \\
\multicolumn{1}{c|}{Transformer} & \multicolumn{1}{c|}{MAE,ires100,fau,DenseNet} & \multicolumn{1}{c|}{V,A,AU} & \multicolumn{1}{c|}{0.5} & 0.5262 \\
\multicolumn{1}{c|}{LSTM}          & \multicolumn{1}{c|}{MAE,ires100,fau,DenseNet} & \multicolumn{1}{c|}{V,AU} & \multicolumn{1}{c|}{0.5}       & 0.5246          \\
\multicolumn{1}{c|}{LSTM}          & \multicolumn{1}{c|}{MAE,ires100,fau,DenseNet} & \multicolumn{1}{c|}{V,AU} & \multicolumn{1}{c|}{0.5}       & 0.5228          \\
\multicolumn{1}{c|}{LSTM}          & \multicolumn{1}{c|}{MAE,ires100,DenseNet}     & \multicolumn{1}{c|}{V,AU} & \multicolumn{1}{c|}{0.5}       & 0.5227          \\ \hline
\multirow{2}{*}{\textbf{Ensemble}} &                                               &                           & 0.5                            & 0.5486          \\
                                   &                                               &                           & \textbf{variable}                       & \textbf{0.5664} \\ \hline
\end{tabular}
\end{table}

Table \ref{tab:AU} shows the results on the validation set for AU prediction. As is shown in the table, the transformer-based model and LSTM-based model can achieve excellent performance for AU and the ensemble result can achieve 0.5664 on the validation set. 
We try two ensemble types for AU. The first is the vote strategy, and we predict 1 when 0 and 1 have the same number of votes. The second is averaging the probabilities from different models for each AU and search the best threshold based on the performance on the validation set for the final prediction. 

\begin{table}[]
\begin{center}
  \caption{The results of the 6-fold cross-validation experiments. The first five folds are from the training set. Fold 6 means the official validation set.}
  \label{tab:5fold}
\begin{tabular}{c|cccc|c}
\hline
         & Valence & Arousal & EXPR   & AU      & $P_{MTL}$   \\ \hline
Fold 1   & 0.6742  & 0.6663  & 0.4013 & 0.5558  & 1.6274     \\
Fold 2   & 0.5681  & 0.6597  & 0.3673 & 0.5496  & 1.5306     \\
Fold 3   & 0.6784  & 0.6536  & 0.3327 & 0.5977  & 1.5963     \\
Fold 4   & 0.6706  & 0.6169  & 0.3851 & 0.5886  & 1.6275     \\
Fold 5   & 0.7015  & 0.6707  & 0.4389 & 0.5409  & 1.6658     \\
Fold 6   & 0.6672  & 0.6290  & 0.4156 & 0.5149  & 1.5786     \\ \hline
Average  & 0.6600  & 0.6494  & 0.3901 & 0.5579  & 1.6027     \\ \hline
\end{tabular}
\end{center}
\end{table}

6-fold cross-validation is also conducted for avoiding overfitting on the validation set. After analyzing the dataset distribution, we find the training set is about five times the size of the validation set, so we divide the training set into five folds, and each fold has approximately the same video number and frame number as the validation set. The validation set can be seen as the 6th fold. 
The feature set \{MAE, ires100, fau, DenseNet\} and the transformer-based structure are chosen for valence, expression and AU prediction. The feature set \{MAE, ires100, fau, DenseNet\} and the LSTM-based structure are chosen for arousal prediction. 
Note that we have features fine-tuned on the s-Aff-Wild2 dataset, which may interfere with the results of the corresponding task, so we remove the features fine-tuned on the s-Aff-Wild2 dataset for corresponding 6-fold cross-validation experiments. 
The results are shown in Table \ref{tab:5fold}. 

\subsection{Results on the test set}

We will briefly explain our submission strategies and show the test results of them, which are demonstrated in table \ref{tab:our_test}.

\begin{table}[]
\begin{center}
  \caption{The results of different submission strategies on the test set.}
  \label{tab:our_test}
 
\begin{tabular}{c|c|c}
\hline
Submit & Strategy      & $P_{MTL}$      \\ \hline
1      & Ensemble 1    & 1.4105         \\
2      & Ensemble 2    & 1.3189         \\
3      & Train-Val-Mix & 1.3717         \\
4      & Ensemble 3    & 1.3453         \\
5      & 6-Fold        & \textbf{1.4361}     \\ \hline
\end{tabular}

\end{center}
\end{table}

We only use a simple strategy for the 1st and 2nd submissions, which means we train models on the training set using the features we extract, and choose the models of best epochs for different tasks.  Specifically, only several models are chosen to ensemble to prevent lowering the result and we use vote strategy for expression and AU ensemble for the 1st submission. Furthermore, more models are used to ensemble and we choose the best ensemble strategy to pursue the highest performance on the validation set for 2nd submission. 

Further, we use two carefully designed strategies for the 3rd and 5th submissions, including Train-Val-Mix and 6-Fold.
Specifically, the Train-Val-Mix strategy means the training and validation set are mixed up for training. In this case, we don't have meaningful validation performance to choose models, so we analyze the distribution of the best epochs for previous experiments under the same parameter setting, and empirically choose the models. The selected epoch interval is from 10 to 19 for valence, from 15 to 19 for arousal, from 15 to 24 for expression, and from 30 to 34 for AU. Further, all these models are used to ensemble for better results.
As for the 6-Fold strategy, five folds are used for the training stage and the rest fold is used for validation each time. Since we get six models under six settings, all six models are used to ensemble to get the final results.
Additionally, the 4th submission is a combination of 2nd and 3rd submissions. 

As is shown in the Table\ref{tab:our_test}, the 6-Fold strategy achieves the best performance on the test set, and the 1st ensemble strategy also achieves competitive performance.

\section{Conclusion}

In this paper, we introduce our framework for the Multi-Task Learning (MTL) Challenge of the 4th Affective Behavior Analysis in-the-wild (ABAW) competition. 
Our method utilizes visual information and uses three different sequential models to capture the sequential information. And we also explore three multi-task framework strategies using the relations of different tasks.
In addition, the smooth method and ensemble strategies are used to get better performance.
Our method achieves the performance of $1.7607$ on the validation dataset and $1.4361$ on the test dataset, ranking first in the MTL Challenge.

\section{Acknowledgement}
This work was supported by the National Key R\&D Program of China (No. 2020AAA0108600) and the National Natural Science Foundation of China (No. 62072462).

%
%
\bibliographystyle{splncs04}
\bibliography{main}
\end{document}